\documentclass[preprint,12pt,authoryear]{elsarticle}

\usepackage[utf8]{inputenc}
\usepackage{amssymb}
\usepackage{graphicx}
\usepackage{mathtools}
\usepackage{amsmath}
\usepackage[dvipsnames,table,xcdraw]{xcolor}
\usepackage{svg}
\usepackage{multirow}
\usepackage{subcaption}
\usepackage{float}
\usepackage{fancyhdr}

\newcommand{\edtwa}{\emph{E-DTWA}}

\journal{Expert Systems with Applications}
\date{September 11, 2022}

\begin{document}

\begin{frontmatter}



\title{Expert Enhanced Dynamic Time Warping Based Anomaly Detection}

\affiliation[inst1]{organization={Department of Cybernetics and Artificial Intelligence, Faculty of Electrical Engineering and Informatics, Technical University of Kosice},
            addressline={Letna 1/9}, 
            city={Kosice},
            postcode={042 00}, 
            country={Slovakia}}
            
\affiliation[inst2]{organization={Kempelen Institute of Intelligent Technologies},
            addressline={Nivy Tower, Mlynske Nivy II. 18890/5}, 
            city={Bratislava},
            postcode={821 09}, 
            country={Slovakia}}

\author[inst1,inst2]{Matej Kloska\corref{cor1}}
\ead{matej.kloska@kinit.sk}
\author[inst2]{Gabriela Grmanova}
\ead{gabriela.grmanova@kinit.sk}
\author[inst2]{Viera Rozinajova}
\ead{viera.rozinajova@kinit.sk}

\cortext[cor1]{Corresponding author}

\begin{abstract}
  Dynamic time warping (DTW) is a well-known algorithm for time series elastic dissimilarity measure. Its ability to deal with non-linear time distortions makes it helpful in variety of data mining tasks. Such a task is also anomaly detection which attempts to reveal unexpected behaviour without false detection alarms. In this paper, we propose a novel anomaly detection method named \emph{Expert enhanced dynamic time warping anomaly detection (E-DTWA)}. It is based on DTW with additional enhancements involving human-in-the-loop concept. The main benefits of our approach comprise efficient detection, flexible retraining based on strong consideration of the expert's detection feedback while retaining low computational and space complexity.
\end{abstract}



\begin{keyword}
Time series \sep Anomaly detection \sep Dynamic time warping.
\end{keyword}

\end{frontmatter}


\thispagestyle{fancy}
\fancyhf{}
\headheight 40pt \headsep 10pt
\renewcommand{\headrulewidth}{0pt}
\lhead{\scriptsize This article has been published in a revised form in \emph{Expert Systems with Applications: An International Journal} [https://doi.org/10.1016/j.eswa.2023.120030]. This version is published under a Creative Commons CC-BY-NC-ND. No commercial re-distribution or re-use allowed. Derivative works cannot be distributed. © Matej Kloska, Gabriela Grmanova, Viera Rozinajova.}

\section{Introduction}
    Nowadays, huge amounts of various data are generated on a daily basis. Intelligent analysis of such data plays a crucial role in decision making in many areas including finance~\citep{anandakrishnan2018anomaly}, security~\citep{mothukuri2021federated}, healthcare~\citep{ukil2016iot}, energy~\citep{malik2021intelligent} and industry~\citep{peres2018idarts}.
    
    When processing these large amounts of data, time and space complexity belongs to the characteristics that must be taken into particular consideration. Special approaches need to be applied in order to make big data analytic tasks computationally tractable.
    Just like the analysis of a large amount of data, it can also be a problem when only a limited amount of data is available. The data sample may not adequately represent the actual distribution of the data in the population and patterns derived from this sample may not be general enough. In fact, small data can even be present in very big datasets when we are interested in better understanding a small sub-sample given by some rare characteristic.
    
    Therefore, anomaly detection is widely considered to be one of the most important part of intelligent data analysis~\citep{cook2019anomaly}. Its task is to observe a system in its environment and point out any kind of distortion or deviation from normal behaviour in incoming data as soon as possible.
    It turns out that behavioural patterns matching is one of the popular methods to detect anomalies~\citep{rettig2019online}, as they allow representing valuable knowledge -- whether expert's, data-mined or obtained by combining those two types of knowledge.  
    
    We focus on anomaly detection in time series. Practically any algorithm for time series anomaly detection makes the decisions based on distances measured between time series. While dealing with ever evolving time series, it is virtually impossible to apply naive, yet powerful, distance measures such as Euclidean Distance (ED). Even small-time distortion results into a great measure penalty. Elastic similarity measures (ESM) are proven to be a promising solution to this challenge as their ability to make flexible alignment between pattern within time series window greatly improves computed similarity measure \citep{lines2015time}.
    
    In the past decades, DTW~\citep{myers1980performance}, as one of ESM representatives, has been widely employed in various machine learning models in domains such as speech recognition~\citep{amin2008speech,juang1984hidden}, signature verification~\citep{wu2019deep}, gesture recognition~\citep{choi2018modified,kowdiki2021automatic}, time series classification~\citep{lahreche2021fast,li2020adaptively} and similarity measure~\citep{choi2020fast} due to its outstanding performance in flexible point-wise alignment. Quadratic complexity of this algorithm makes it hard to utilise in online processing. However, extensive research has been conducted in the area of computational optimisation with encouraging results~\citep{itakura1975minimum,rakthanmanon2012searching,sakoe1978dynamic}.
    
    While solving data analysis tasks such as anomaly detection, it is often beneficial to enhance knowledge extracted by machine learning models with human knowledge. Human-in-the-loop is a paradigm, in which human knowledge is combined with machine learning models in a continuous loop to solve problems more effectively and accurately. Human-in-the-loop anomaly detection is defined by \citet{calikus2022together} as a model which incorporates a different form of human feedback into the detection process to achieve more robust and effective anomaly detection. A domain expert may help to label the observations that are considered to be anomalous by the algorithm or those which the algorithm is not confident with. This process is referred to as active learning~\citep{settles2009active} and its aim is to select the most informative observations to be labelled by a human expert. By retraining the model based on human labelling, the anomaly detection algorithm becomes more accurate in the long-term.
    
    Our goal is to design an effective anomaly detection method for time series analysis with low computational complexity and the ambition to be applicable to both large and small datasets. We propose a method based on elastic pattern matching, allowing a human to intervene into the detection process and offer relevant feedback. The reason for focusing on elastic similarity measures lies in their ability to compensate time distortions and to compare similar processes proceeding in variable pace. The application of human-in-the-loop concept enables us to use the model in situations when we cannot rely only on knowledge extracted from limited datasets.
    
    This paper is organised as follows. Section~\ref{sec:related_work} describes the original dynamic time warping method, anomaly detection techniques and human-cen\-tered artificial intelligence approach as well as the state-of-the-art in anomaly detection methods utilising these approaches. Section~\ref{sec:method} introduces \edtwa~- our method for anomaly detection. Section~\ref{sec:evaluation} deals with an experimental evaluation of our method on datasets from three different domains. Lastly, section~\ref{sec:conclusion} offers conclusions and suggestions for future work.
    
\section{Related work}
    \label{sec:related_work}

    This section presents related work on anomaly detection, outlines the human-in-the-loop concept, and explains the need for a new complex method for time series anomaly detection.
    
    \subsection{Anomaly detection}
        The aim of anomaly detection is to detect whether a newly observed point or motif is novel or normal, when comparing it to a set of normal behaviours~\citep{chandola2009anomaly}.
        
        Anomaly detection can be addressed as supervised, unsupervised, or semi-supervised problem, mainly with regard to the character of available data. Having labelled set with both normal and anomalous samples, supervised classification task can be formulated and solved based on many different approaches, such as nearest neighbour, decision trees or neural networks. When no labelled data are available, solving unsupervised clustering task is the only way. Semi-supervised approaches train one-class models such as support vector machines~\citep{Montague2017efficient} or neural networks~\citep{Akcay2018GANomaly} and train them only on the set of normal data samples. 
        
        \emph{Local Outlier Factor} (LOF)~\citep{breunig2000lof} as an example of nearest neighbour based methods tries to represent object being anomaly at some degree regarding to its neighbourhood surrounding rather than strictly assign either normal or abnormal class.
        LOF method computes the local density of a given sample with respect to its neighbours. The samples with a substantially lower density than their neighbours are considered to be anomalies. Computational complexity of such a method is $O(n^2)$, where $n$ is number of objects.
        LOF-based methods are usually based (as the former one) on the nearest neighbour analysis~\citep{tang2002enhancing,papadimitriou2003loci} as-well-as clustering~\citep{he2003discovering,amer2012nearest}.
        
        \emph{Isolation forest} (IF) as another detection method with its extensions was successfully used in the anomaly detection areas such as Machine Monitoring Data~\citep{li2021similarity} and streaming data~\citep{togbe2020anomaly}. The method tries to isolate observed samples by selecting randomly any from available features and then randomly selecting threshold value between the maximum and minimum values of the selected feature. With recursive partitioning application, we can produce tree structure. Anomalies have usually significantly shorter path from the root node to the leaf node. Computational complexity of such a method is $O(\operatorname{n}\operatorname{log}n)$, where $n$ is number of objects. Recently, extended version~\citep{hariri2019extended} of the method was proposed with improved robustness.
        
        In recent years, interest in anomaly detection using neural networks has grown. Neural networks based detection plays important role in different areas such as monitoring~\citep{naseer2018enhanced}, industrial technologies monitoring~\citep{su2019robust,zhou2020siamese}, healthcare~\citep{pereira2019learning} or financials~\citep{hilal2022financial}. The major drawback of neural networks based anomaly detection lies in their black-box principle --- it is hard to validate the detection process by an expert.
        
        To pose the problem of anomaly detection in any detection system implies the existence of a subjacent concept of normality \citep{zhang2009survey}. Having the normal model mostly trained in a semi-supervised manner (normality concept), the task of the detection is to find a degree to what unseen sample matches the normal model --- higher match result implies lower chance for the sample to be anomalous.

        In many situations, it is difficult to match normal behaviour to observed time series by naive similarity measures such as ED. This approach often results in high false positives detection. DTW appears to be a viable solution~\citep{teng2010anomaly,jones2016exemplar,diab2019anomaly}, whereas it provides elasticity necessary to reach the best possible alignment --- lowering false positives while detecting anomalies.
        
        Anomaly detection use case based on normal behaviour is mostly useful for monitoring applications where incoming data is generated from the same data distribution with minimum changes in stationary interval. From real-life applications, it is nearly impossible to fulfil the given precondition because change in stationary interval is usually related to concept drift. Decision whether it is just temporal (anomaly) or gradual change in generator (concept drift) is challenging, thus, expert knowledge is sometimes necessary to review detected change for anomalous behaviour.
        
        One of methods that try to describe training time series with expectable normal behaviour and consequent matching is Exemplar learning ~\citep{jones2016exemplar}. The method was proposed to describe time series as feature vectors that capture both the high frequency and low frequency information in sets of similar subsequences of the time series. According to the results, this approach is more efficient than brute force algorithms but lacks the ability to incorporate expert knowledge.
        
    \subsection{Human-in-the-loop}
\begin{figure}
        \centering
        \includegraphics[width=0.8\linewidth]{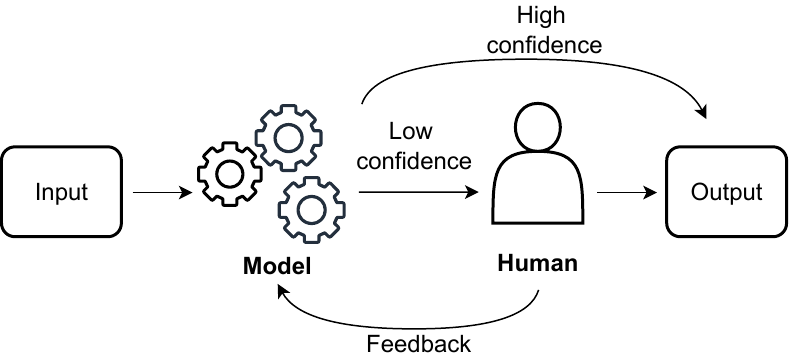}
            
        \caption{Human-in-the-loop: active learning schema. The aim of active learning is to use feedback from the expert for the observations where the algorithm has low confidence of the outcome and to use this feedback to retrain the model.}
        \label{fig:hitl_loop}
    \end{figure}

        Machine learning (ML) methods are able to solve many specific tasks with high accuracy in lab settings. However, the decision whether the ML method is applied in real life depends not only on its potential benefits but also on the risks. Even the best method learned on corrupted or incomplete data can be biased and cause great damage. Thus, in domains such as medicine or security, where the cost of false decision is high, it is desirable not to replace but enhance the decisions of experienced experts in order to take the advantage of both human and ML intelligence. This concept is referred to as human-centered artificial intelligence (HCAI) and represents the main concept of the third wave of artificial intelligence~\citep{xu2019perspective}. More specifically, human-in-the-loop (HITL) is the term defining a set of strategies for combining human and machine intelligence~\citep{monarch2021human}. Human-in-the-loop design strategies can often improve the performance of the system compared to fully automated approaches or humans on their own solutions. The expert intervenes in the ML process by solving tasks that are hard for computers~\citep{wu2022survey}. They provide a-priori knowledge, validate the results, or change some aspect of the ML process. The human intervention continues in a loop in which the ML model is trained and tuned. With each loop, the model becomes more confident and accurate. In active learning (AL), a branch of HITL, a subset of observations is asked to be labelled by humans. AL methods aim to iteratively seek the most informative observations~\citep{budd2021survey}, typically the observations where the algorithm has low confidence of the outcome (see Figure~\ref{fig:hitl_loop}). 
        
        As with other ML tasks, there is no universal ML method for anomaly detection that is applicable for all domains~\citep{freeman2019human}. It is necessary to capture the specifics of the given domain and adapt the anomaly detection process to it. Thus, the application of the HITL concept may not only significantly improve the results of the ML method for anomaly detection but also enable it to be practically deployed even in domains with high cost of errors. 
    
    There were many HITL and especially AL methods for anomaly detection presented in the literature. A nearest neighbor method was applied in~\citep{he2007nearest}. An ensemble of unsupervised outlier detection methods was combined with expert feedback and a supervised learning algorithm in~\citep{veeramachaneni2016ai}. Research in~\citep{chai2020human} applied a clustering method to detect observations that are unlikely outliers and used bipartite graph-based question selection strategy to minimize the interactions between ML and the expert. Active Anomaly Discovery (AAD) algorithm was proposed to incorporate expert provided labels into anomaly detection~\citep{das2016incorporating,das2017incorporating}. AAD selects a potential anomaly and presents it to the expert to label it as a nominal data point or as an anomaly. Based on user feedback, the internal model is updated. This process is repeated in the loop until a given number of queries is spent.
    
    More specifically, AL was also applied for anomaly detection in time series data. Anomaly detection framework for time series data named iRRCF-Active was proposed in~\citep{wang2020practical}. The framework combined an unsupervised anomaly detector based on Robust Random Cut Forest, and an AL component. Semi-supervised approach for time series anomaly detection combining deep reinforcement learning and AL was presented in~\citep{wu2021rlad}. In~\citep{bodor2022little}, the feedback from experts was obtained through AL in order to verify the anomalies. The representative samples to be queried were selected by a combination of three query selection strategies. 
    
    All presented approaches justify the application of the HITL approach in the anomaly detection domain and present significant improvements in comparison to the pure ML approaches without any human intervention. 

    \subsection{Summary}
    A large number of time series anomaly detection methods is presented in the literature. Some of them are characterised by high performance \citep{schlegl2019f}, others by low detection complexity (e.g. neural network based methods~\citep{hilal2022financial,zhou2020siamese}). However, there is a lack of methods that can combine all three advantages simultaneously:
    \begin{itemize}
        \item achieve performance of the state-of-the-art methods, 
        \item do not suffer by high time and space detection complexity,
        \item allow the involvement of an expert to improve the model performance by retraining the model based on expert evaluation.
    \end{itemize}
    
\section{Expert enhanced dynamic time warping based anomaly detection}
    \label{sec:method}
    In this section, we propose novel efficient anomaly detection method based on well-known dynamic time warping. The method is able to reveal the anomalies in the processed time series, incorporate expert knowledge to increase accuracy and adapt to independently detected concept drift while preserving computational complexity at acceptable level. Our method builds up on several premises:
    \begin{itemize}
        \item Elastic similarity measures are able to measure the similarity between time series evolving at different speeds by searching for the best alignment. \citep{oregi2019line}
        
        \item Representation of training set of normal time series by a pair consisting of a \emph{representative time series} and a \emph{matrix of warping paths} between the representative and all training time series can be used to effectively find the anomalies.
        
        \item DTW between normal behaviour pattern and unseen normal time series results in warping path following the diagonal in distance matrix. Based on this observation, reduction of complexity of DTW computation can be proposed. 
        
        \item Building up universal detection model without domain knowledge is nearly impossible and thus incorporating the expert knowledge into the model leads to more accurate results.
    \end{itemize}
    The following sections outline the ideas necessary to describe our novel method \emph{Expert Enhanced Dynamic Time Warping Anomaly Detection (\edtwa)}. It starts with description of DTW method in Section \ref{chapter:dtw} that is obviously not our contribution but is crucial for understanding the method proposed in Sections \ref{edtwa:warping_paths_probabilities}-\ref{model_architecture}.
    
    \subsection{Dynamic Time Warping}
        \label{chapter:dtw}
        DTW similarity measure~\citep{myers1980performance} is commonly used to measure similarities between two temporal sequences. It is heavily utilised in the situations where time distortions occur (e.g. different speech speed or gesture dynamics) and comparison is not trivial or is leading to wrong results when used with naive measures such as Euclidean Distance. DTW can be also applied to measure similarity of sequences with different lengths.
        
        An increasing number of studies have applied DTW for different tasks such as time series classification~\citep{jeong2011weighted,kate2016using,petitjean2014dynamic}, time series indexing~\citep{rakthanmanon2012searching,tan2017indexing}, online signature recognition~\citep{faundez2007line}, anomaly detection~\citep{diab2019anomaly} or pattern matching~\citep{berndt1994using,adwan2012improving}, proving DTW as widely utilised similarity measure.
        
\begin{figure}
            \centering
            \fbox{\includegraphics[width=0.75\linewidth, height=80pt]{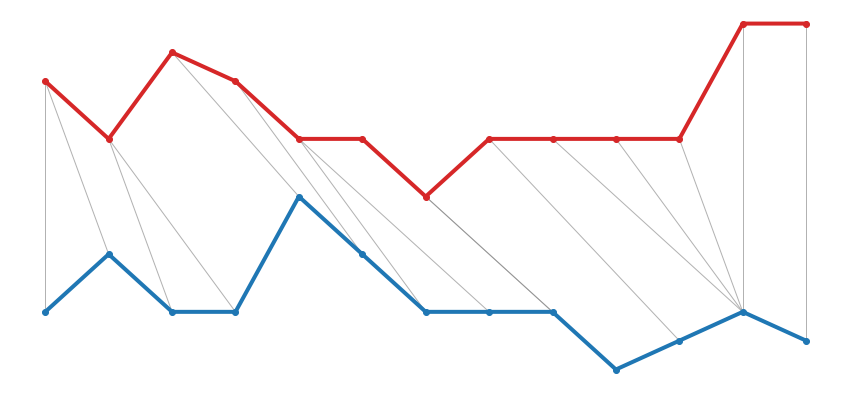}}
            \caption{An example of minimum point-wise alignment between time series A (red) and B (blue). Concept of best alignment leads to mapping first point on time series A to 2 different points on time series B.}
            \label{fig:dtw_alignment}
        \end{figure}

        Formally, DTW measure between two time series $X^m = (x_1, x_2, ..., x_m)$ and $Y^n = (y_1, y_2, ..., y_n)$ is given by the minimum cumulative point-wise alignment between both time series. Fig. \ref{fig:dtw_alignment} depicts an example of alignment between two given time series.
        
        The alignment between two time series $X^m$ and $Y^n$ is represented by a path $p = \{(i_1, j_1), ..., (i_m, j_n)\}$ starting at point $(1, 1)$, reaching point $(m, n)$ in $[1; m] \times [1, n]$ matrix lattice. Each point $(i, j)$ represents alignment of the points $x_i$ and $y_j$. Any path $p$ fulfilling point-wise condition $(i_m, j_n) - (i_{m-1}, j_{n-1}) \in \{(1,0), (1,1), (0,1)\}$, i.e., allowing only movements in $\rightarrow, \nearrow$ and $\uparrow$ directions, is considered to be $allowed$. Set $P$ contains all allowed paths between time series $X^m$ and $Y^n$ in the $[1, m] \times [1, n]$ matrix lattice.
        
        To fully define DTW measure, we need to define path weight $w(p)$ as:
        
        \begin{equation}
            w(p) = \sum_{(i,j) \in p}d_{i,j},
        \end{equation}
        where $d_{i, j} = |x_i - y_j|$ is point-wise distance returning values in range $\left[ 0; \infty\right)$.
        
\begin{figure}
            \centering
            \includegraphics[width=0.45\linewidth]{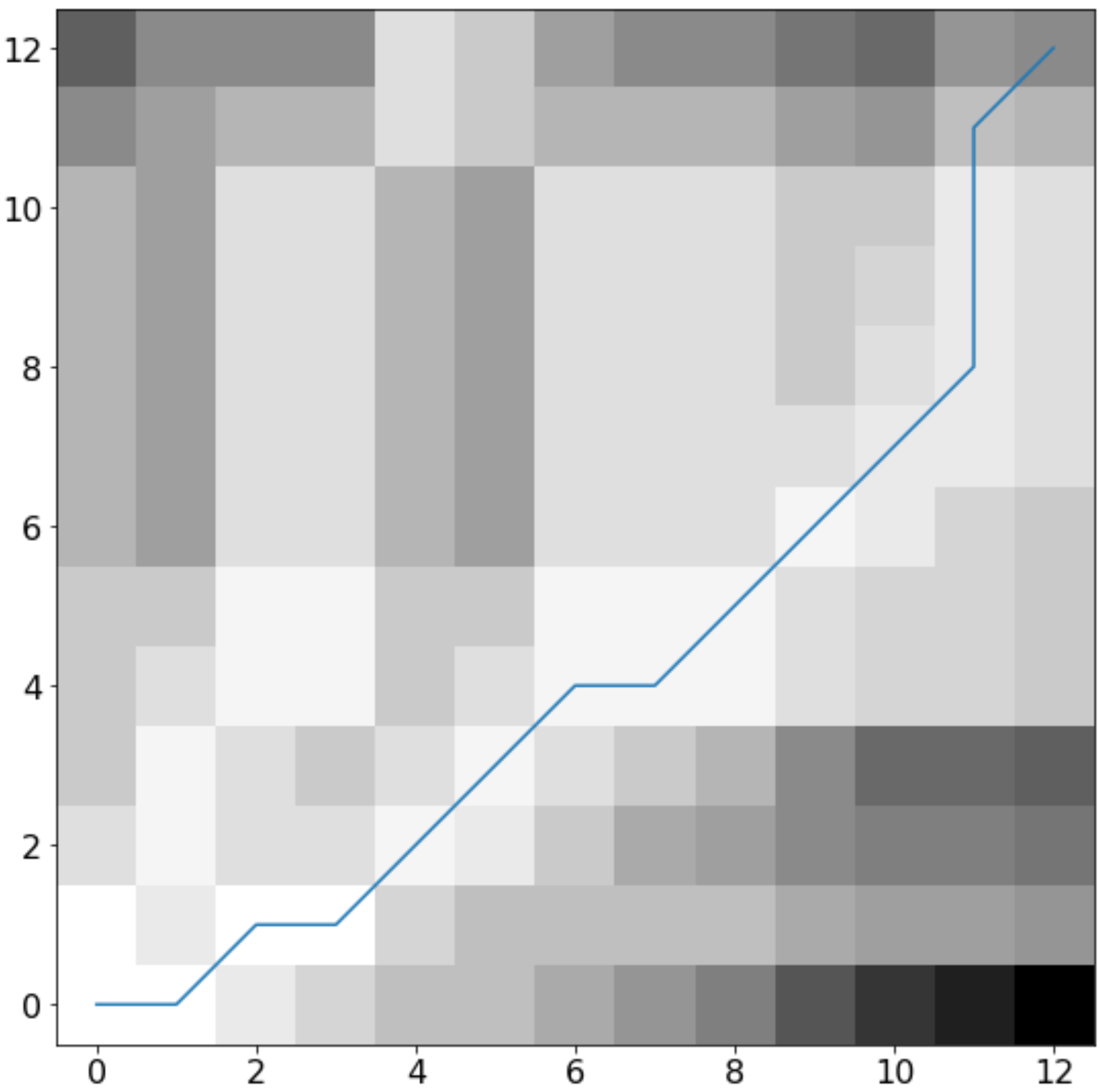}
            \caption{Minimum warping path (blue line) spanning from left bottom corner (0,0) to top right corner (12,12). Axes X and Y represent time series of length 13. Gray scaled positions in the matrix represent alignment distance, i.e., white positions have the lowest alignment distance while black have the highest.}
            \label{fig:dtw_warping_path}
        \end{figure}

        DTW measure between time series $X^m$ and $Y^n$ following already stated constraints is defined as:
        
        \begin{equation}
            D(X^m, Y^n) = \min_{p \in P}(w(p))
        \end{equation}
        From this point, we will denote DTW distance between time series $X^m$ and $Y^n$ as $D_{m,n}$ and the optimal (and also minimal) warping path associated with it as $p^{X,Y}$. Fig.~\ref{fig:dtw_warping_path} shows minimal warping path between time series from Fig.~\ref{fig:dtw_alignment}.
        
        With increasing length of time series $X^m$ and $Y^n$ the number of allowed paths in set $P$ is rapidly growing, leading in exhaustive computational searching of optimal path. To address this problem, we can employ dynamic programming with recursion defined as:
        \begin{equation}
            D_{m,n} = d_{m,n} + min(D_{m-1, n}, D_{m-1, n-1}, D_{m, n-1}),
            \label{eq:dtw}
        \end{equation}
        with initial cases: $D_{0,0} = 0$, $D_{i,0} = \infty$ for $i \in \{1, ..., m\}$ and $D_{0,j} = \infty$ for $j \in \{1, ..., n\}$.
        Computational complexity of this recursive algorithm is $\mathcal{O}(m \cdot n)$, making this algorithm unfeasible for long input time series. To address this issue, the search space needs to be reduced by applying additional path constraints. Multiple similar approaches including Itakura parallelogram~\citep{itakura1975minimum}, Sakoe-Chiba band~\citep{sakoe1978dynamic} and Ratanamahatana-Keogh (R-K) band~\citep{ratanamahatana2004making} were proposed to reduce the computational complexity to an acceptable level. 
    
    \subsection{Normal time series representation}
        \label{edtwa:warping_paths_probabilities}
        Given training set $\mathcal{T} = \{T_1, T_2, ..., T_N\}$ of time series with normal behaviour, a \emph{representative} normal time series $R$ is extracted such that it represents a typical normal behaviour pattern\footnote{If there are several typical normal behaviour patterns, training set is split into several subsets and each subset is processed separately.}. This can be done either by an expert or the representative might be pattern-mined from training data. 
        Training set $\mathcal{T}$ is then represented by a pair $(R, \mathbf{M})$, where $R$ is the representative time series and  $\mathbf{M}$ is \emph{warping matrix} that is built up of $N$ warping paths between $R$ and $T_1, T_2, ..., T_N$. To build up the warping matrix $\mathbf{M}$, a~set of warping paths $\mathcal{W} = \{{p^{R, T} \hspace{2pt} \vert \hspace{2pt} T \in \mathcal{T}\}}$ is firstly created by DTW
        for pairs of representative normal pattern $R$ and all training time series $T \in \mathcal{T}$.
        
        Secondly, we define an encoding function $\operatorname{enc}(p_i)$ which is applied to warping path step position $i, i \in \{1, ..., n\}$ and based on $p_{i}$ and $p_{i-1}$ it returns a binary 3-dimensional vector:
        \begin{equation}
            \resizebox{0.91\hsize}{!}{
                $\operatorname{enc}(p_i) =
                    \begin{cases}
                    (1, 0, 0) & \text{: $\operatorname{col}(p_i)-1 = \operatorname{col}(p_{i-1}) \wedge \operatorname{row}(p_i) = \operatorname{row}(p_{i-1})$,} \\
                    (0, 1, 0) & \text{: $\operatorname{col}(p_i)-1 = \operatorname{col}(p_{i-1}) \wedge \operatorname{row}(p_i) - 1 = 
                \operatorname{row}(p_{i-1}) $,} \\
                    (0, 0, 1) & \text{: $\operatorname{col}(p_i) = \operatorname{col}(p_{i-1}) \wedge \operatorname{row}(p_i) - 1 = \operatorname{row}(p_{i-1})$,} \\
                    (0, 0, 0)& \text{: otherwise.}
                    \end{cases},
                $
            }
        \end{equation}
        where $p_i$ denotes $i-th$ step in warping path $p$, $col(p_i)$ and $row(p_i)$ functions return row and column index in distance matrix based on provided warping path step respectively. The returned binary vector carries information about the previous step direction leading towards current path position $p_i$; respectively in a non-descent direction $\rightarrow, \nearrow$ and $\uparrow$ to above mentioned function cases.
        
        Finally, having a set of normal warping paths $\mathcal{W}$, we construct $m \times n$ \emph{warping matrix} $\mathbf{M}$ (Fig. \ref{fig:edtwa_occurences_matrix})
        such that:
         \begin{equation}
            \label{eq:edtwa:probabilities_matrix}
            M_{k,l} = \sum_{p_i\in \mathcal{W}: \operatorname{row}(p_i) = k\  \wedge~\operatorname{col}(p_i) = l}  \operatorname{enc}(p_i).
        \end{equation}
       
        \begin{figure}
            \centering
            \begin{equation*}
                \resizebox{\hsize}{!}{
                    $
                    \begin{aligned}
                        \begin{array}{rcl}
                         &  & \{((0,0), (1,1), (2,2), (3,3)), \\
                                    &   & ( (0,0), (0,1), (1,2), (2,3), (3,3)), \\
                        \mathcal{W}            &  = & ((0,0), (1,1), (2,1), (3,2), (3,3)), \\
                                    &   & ((0,0), (1,1), (2,1), (3,2), (3,3)), \\
                                    &   & ((0,0), (0,1), (1,2), (2,3), (3,3))\}
                        \end{array}
                         \quad
                        \mathbf{M}= \begin{bmatrix}
                        (0, 0, 0) & (0, 0, 0) & (0, 2, 0) & (2, 1, 2) \\
                        (0, 0, 0) & (0, 0, 2) & (0, 1, 0) & (0, 2, 0) \\
                        (0, 0, 0) & (0, 3, 0) & (0, 2, 0) & (0, 0, 0) \\
                        (0, 0, 0) & (2, 0, 0) & (0, 0, 0) & (0, 0, 0)
                        \end{bmatrix}
                    \end{aligned}
                    $
                }
            \end{equation*}
            \caption{Example of constructed $M^{4,4}$ matrix based on calculated paths from set $\mathcal{W}$.}
            \label{fig:edtwa_occurences_matrix}
        \end{figure}
        The matrix $\mathbf{M}$ thus captures deviations of training normal time series from representative normal pattern and represents initial warping matrix that can be updated in a loop by interaction with human expert. 
        
\begin{figure}
            \centering
             \subfloat[\centering Single behaviour data]{{\includegraphics[width=0.45\linewidth]{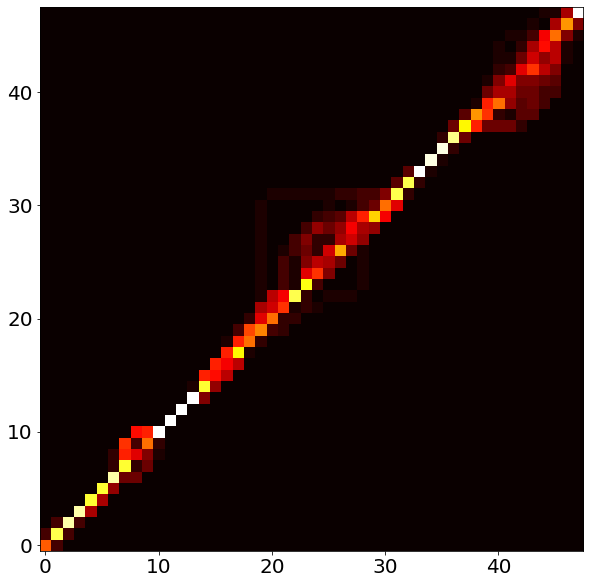}}}%
            \qquad
            \subfloat[\centering Mixed behaviour data ]{{\includegraphics[width=0.45\linewidth]{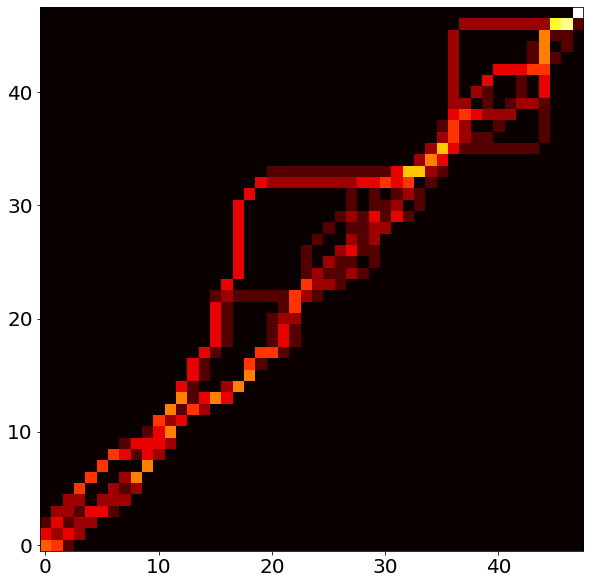} }}%
            
            \caption{Examples of two warping matrices $\mathbf{M}$ visualised as heatmap. Brighter areas in the matrix represent locations with significantly higher probability of warping path. Example a) shows correctly selected representative normal pattern and matching data case. Example b) shows a case where provided training data are mixed from multiple behaviours - should be split into multiple patterns.}
            \label{fig:edtwa_probabilities_heatmap}
        \end{figure}

        Fig. \ref{fig:edtwa_probabilities_heatmap}. shows an example of $48 \times 48$ warping matrix $\mathbf{M}$. Bright areas in the figure reflect high incidence of normal warping paths. Such a visual representation helps us to validate whether a given normal pattern and training data match together. If the bright area closely matches the diagonal, representative normal pattern matches the data and it is suitable for the detection (example~a); if not, we need to adjust either the pattern or split the data and introduce more normal patterns (example~b).

        \subsection{Detection of unseen time series}
        Given an unseen time series $Q$, by DTW method we calculate a warping path $p^{R,Q}$ 
        with normal pattern $R$. To detect whether $Q$ is anomaly, the whole warping path is inspected in overlapping path parts (windows) of length $l$ for how much they are supported by warping paths in $\mathbf{M}$. The path parts $(p_{i-l}^{R,Q}, \dots, p_{i-1}^{R,Q})$ are defined by their length $l$ end point $p^{R,Q}_i$. 
        As says the proverb ``A chain is only as strong as its weakest link'', the support of path part is given by the support of the least supported point of that path part. The support of path part is formally defined as:
        \begin{equation}
            supp(p^{R,Q}_i, l) = 
                \max_{j \in [i-1..i-l]} \{enc(p^{R,Q}_j) \cdot M_{col(p^{R,Q}_j),row(p^{R,Q}_j)}\}.
        \end{equation}

        The relative support is then computed by 
        \begin{equation}
            rsupp(p^{R,Q}_i, l) = \frac{
                supp(p^{R,Q}_i, l)
            }{
                count\_paths(p_i^{R,Q})
            },
        \end{equation}
        where $l$ $(1 < l \ll \min(|R|,|Q|))$ is a parameter denoting a width of the inspected path part and the function $count\_paths(p_i)$ returns the number of all paths from $\mathbf{M}$ passing through $p_i^{R,Q}$ and is given by:
        \begin{equation}
            count\_paths(p_i)=\sum_{k = 1}^{3}(M_{col(p_i),row(p_i)})_k.
        \end{equation} 
        Figure \ref{fig:edtwa_probability_calculation} presents an example of relative support calculation.
        
        \begin{figure}
            \centering
            \begin{equation*}
                \resizebox{.9\hsize}{!}{
                    $
                    \begin{aligned}
                        p^{R,Q} = \begin{bmatrix}
                        0 & 0 & 0 & \textbf{1} \\
                        0 & 0 & 0 & \textbf{1} \\
                        0 & 0 & \textbf{1} & 0 \\
                        1 & 1 & 0 & 0 \\
                        \end{bmatrix}
                        , \quad
                        \mathbf{M}= \begin{bmatrix}
                        (0, 0, 0) & (0, 0, 0) & (0, 2, 0) & \textbf{(2, 1, 2)} \\
                        (0, 0, 0) & (0, 0, 2) & (0, 1, 0) & \textbf{(0, 2, 0)} \\
                        (0, 0, 0) & (0, 3, 0) & \textbf{(0, 2, 0)} & (0, 0, 0) \\
                        (0, 0, 0) & (2, 0, 0) & (0, 0, 0) & (0, 0, 0)
                        \end{bmatrix}
                    \end{aligned}
                    $
                }
            \end{equation*}
            \caption{Example of $rsupp(p^{R,Q}_5, 2)$ calculation for path parts of length $l=2$ under constructed warping path $p^{R,Q}$ transformed into binary matrix and occurrences matrix $\mathbf{M}$. Result $rsupp(p^{R,Q}_5,2) = 0.4$}
            \label{fig:edtwa_probability_calculation}
        \end{figure}
        
        To detect whether path part given by length $l$ and end-point $p_i$ is normal vs. anomalous, we define function $detect(p_i,l)$ as:
        \begin{equation}
            detect(p_i, l) = \begin{cases}
                1 :  count\_paths(p^{R,Q}_i) > 0~\wedge \\ ~~~~rsupp(p_i, l) \geq \theta(p_i,l, \mathbf{M}) \\
                0 : \text{ otherwise}.
            \end{cases}
        \end{equation}
        \sloppy Function compares $rsupp(p^{R,Q}_i,l)$ to the training data specific threshold $\theta(p^{R,Q}_i,l, \mathbf{M})$. The threshold is computed as the ratio between minimal $rsupp()$ of all path parts of length $l$ in $\mathbf{M}$ passing through $p^{R,Q}_i$ in the same direction and the number of total path parts of length $l$ in $\mathbf{M}$ passing through the position $p^{R,Q}_i$.
        
        Based on the $detect(p_i, l)$ function, we define \edtwa~normality detection score expressing the degree of similarity of time series $Q$ to the set of training time series $\mathcal{T}$ (represented by pair $(R, \mathbf{M})$) by:
        \begin{equation}
            EDTWA_{R,\mathbf{M}, l}(Q) = \frac{
                \sum_{i=1}^{|p^{R,Q}|}detect(p^{R,Q}_i, l)
            }{
                \vert p^{R,Q} \vert
            },
        \end{equation}
        
        Intuition behind the \edtwa~method is to express normality detection score as a ratio of the number of detected normal path parts and the total number of path parts. Total number of path parts is given by the number of path steps defining end-points and is given by the length of warping path $|p^{R,Q}|$. 

        The method returns normal certainty score in the range $\left[0;1\right]$, meaning certainly anomalous time series (0) and certainly normal time series (1).
        
        Given \edtwa~return value, there are two options how to define anomalous threshold $\theta_{edtwa}$ for in-between values:
        \begin{itemize}
            \item \textit{expert-based}: an expert is involved in threshold definition based on calculated training relative support scores,
            \item \textit{data-driven}: the lowest gained relative support score on training data is automatically declared as threshold.
        \end{itemize}
        
    \subsection{Reduction of the number of paths and model complexity}
        \label{edtwa:reduction_and_complexity}
        As discussed in Section~\ref{chapter:dtw}, quadratic complexity of DTW is a crucial problem with common solution based on introducing path constraints, narrowing down the space of allowed paths to the area around the diagonal in a distance matrix. Either we can exploit family of "band" base constraints~\citep{itakura1975minimum,sakoe1978dynamic} or use more general approach based on calculating DTW on reduced time series with consequent narrowing down space and increasing details~\citep{ratanamahatana2004making}. In this chapter, we propose a simple, yet powerful, method for asymmetric constraining distance measure matrix calculation.
        
        Given a set of warping paths $\mathcal{W}$ we construct $n \times m$ binary matrix $\mathbf{N}$ such~as:
        \begin{equation}
            N_{i,j} = \begin{cases}
                1 \text{ : $path\_step\_exists_{p \in \mathcal{W}}(p_{i,j})$} ,\\
                0 \text{ : otherwise}
            \end{cases}
        \end{equation}
        where $path\_step\_exists()$ function returns boolean value indicating whether given position exists on the given path.
        
        Applying this binary matrix as input mask to allow (1) and forbid (0) warping path calculation for arbitrary distance matrix position, we effectively limit DTW calculation only to positions, where covering normal warping paths is highly probable. This idea is more general, purely data-driven and it is inspired by Ratanamahatana band~\citep{ratanamahatana2004making}.
        
        There are two different warping path cases we need to consider for validation:
        \begin{itemize}
            \item \textit{fully inside constrained region:} warping is covered inside region, there is no significant shift out of region --- behaviour is covered by $\mathcal{W}$,
            \item \textit{possible leaning outside region:} matched time series contains significant distortions leanings outside of constrained region --- warping path lies along region borderline with significantly higher distance measure at the end.
        \end{itemize}

        Computational complexity of \edtwa~score calculation and hence the detection process of \edtwa~method consists of DTW and $detect(p_i)$ computational complexities. DTW can be computed in $\mathcal{O}(l \cdot max\{m,n\})$ where $l$ is maximum width of allowed region within binary matrix $\mathbf{N}$ and $m,n$ are lengths of $R$ and $Q$. Complexity of function $detect(p_i)$ is linear and depends on the maximum warping path length with worst (but highly improbable) case $\mathcal{O}(m + n)$.
        
        Computational complexity of an unseen time series detection for \edtwa~ is $O(k \cdot l \cdot max\{m,n\})$ and $LOF$ is $O(d \cdot l \cdot max\{m,n\})$ where $k$ is number of normal patterns and $d$ is number of neighbours. At worst case, $d$ is equal to the number of time series in the dataset, hence $k << d$.
    
    \subsection{Method architecture and the model update}
        \label{model_architecture}
\begin{figure*}
            \centering
            \includegraphics[width=0.9\linewidth]{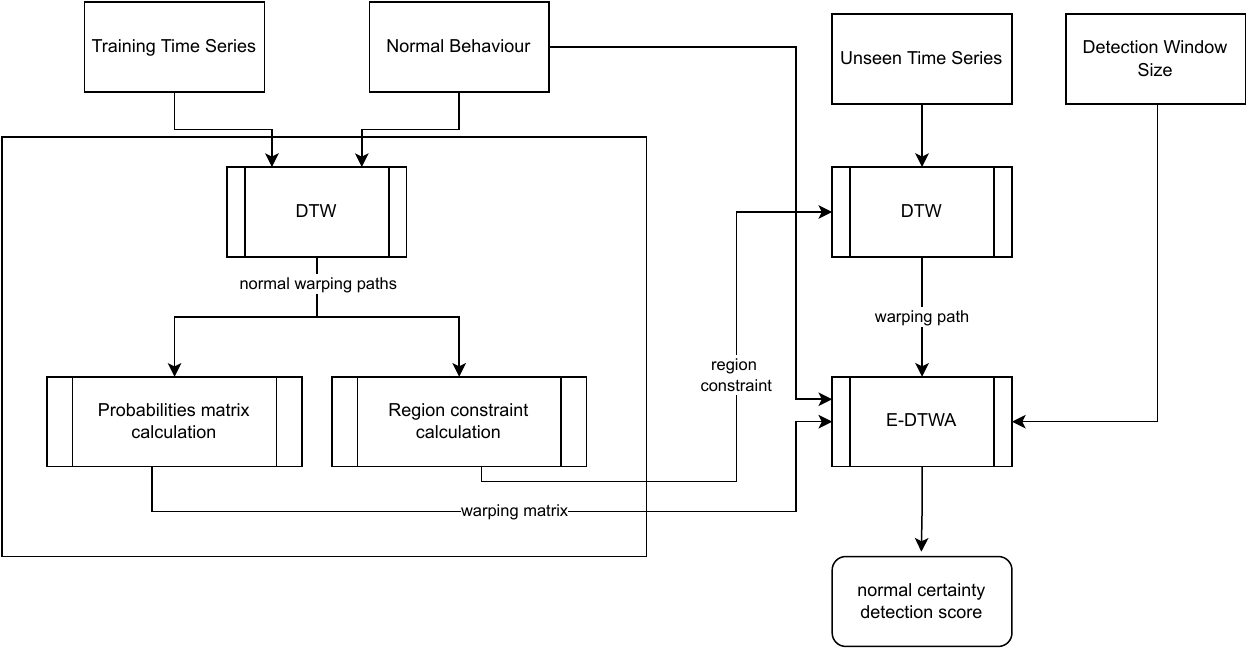}
            
            \caption{The flow of \edtwa~method. Left top-down flow depicts training phase of our method with necessary inputs at the top. Results of the training (region constraint and probabilities matrix) are passed into detection phase (right top-down flow). The result is returned as normal detection score (bottom right corner).}
            \label{fig:edtwa_architecture}
        \end{figure*}

        Overall method architecture in the context of training and detection phases with all necessary dependencies is depicted in Figure \ref{fig:edtwa_architecture}. Note, that model is trained on a semi-supervised manner, where representative normal patterns together with warping matrix represent the normality concept.
        
        However, in anomaly detection, two key aspects need to be addressed: feedback loop adoption, and concept-drift adoption. Both aspects lean towards efficient way to adopt incoming changes into detection model without any need to retrain from scratch. Given existing warping matrix $\mathbf{M}$ and expert's feedback for time series $Q$ that is normal, we enrich detection model by matched time series $Q$ by:
        
        \begin{equation}
            \label{eq:edtwa:probabilities_matrix_update}
            M_{k,l}^{new} = M_{k,l}^{old} + \sum_{p_i^{R,Q}: \operatorname{row}(p_i^{R,Q}) = k~\wedge~\operatorname{col}(p_i^{R,Q}) = l}  \operatorname{enc}(p_i).
        \end{equation}
        
        With expert's anomalous time series feedback, we handle update as reverse operation, i.e., subtraction. We apply step-wise path vector subtraction in equation \ref{eq:edtwa:probabilities_matrix_update} resulting in lower warping path support. Allowing learning from anomalous samples turns the semi-supervised concept of the method into the supervised one.
        
        The benefit of such an update method is in low and constant memory requirements. Update does not require any additional memory to copy warping matrix, with proper concurrency coordination, therefore, it is possible to perform quick update.
        
        We can apply the same approach for batch model changes too --- forgetting large portions of the past trained time series and including newly discovered trends --- our method thus provides a support for concept drift issues. However, the concept of drift detection itself is a complex task and it is not directly the concern of this paper.
    
\begin{figure*}[!ht]
            \centering
            \includegraphics[width=0.75\linewidth]{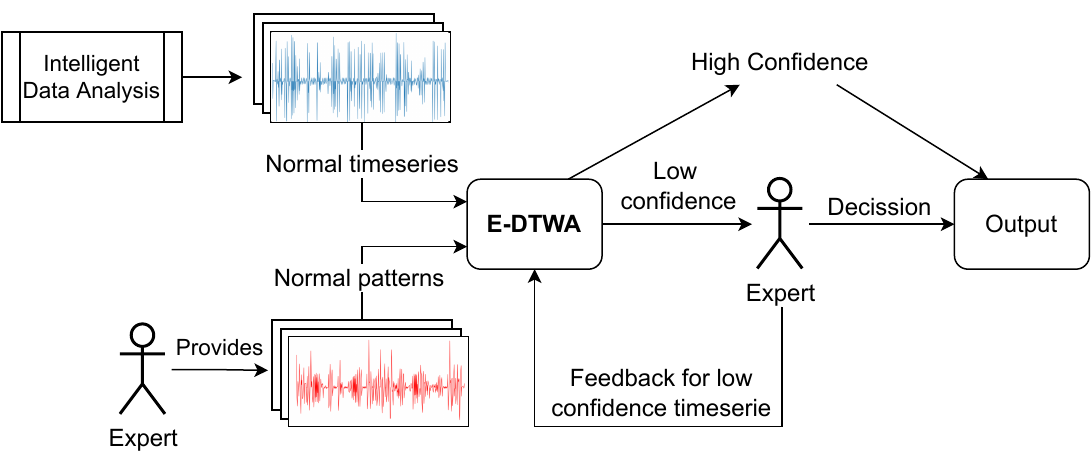}
            
            \caption{Role of \edtwa~method from the \emph{HITL} concept point of view. The schema also shows off the role of expert at the input --- normal patterns provisioning as well as at active learning phase --- low confidence decisions.}
            \label{fig:edtwa_hitl}
        \end{figure*}

        Figure~\ref{fig:edtwa_hitl}. depicts \edtwa~method role within HITL concept as well as the expert knowledge support.
    
\section{Evaluation}
    \label{sec:evaluation}
    In Section \ref{edtwa:reduction_and_complexity}, we showed that the proposed method belong among methods with the smallest detection space and time complexity. To evaluate the other characteristics of the method, we show in the following experiments that:
    \begin{enumerate}
        \item the performance of the proposed method is comparable to the best performances of the considered methods,
        \item the expert knowledge incorporation for the model retraining increases the performance of the method.
    \end{enumerate}
    
    \subsection{Performance Evaluation}
    To evaluate our method, we compare anomaly detection performance with simple baseline approach based on DTW weighted path ($DTW_{base}$) and two well-known methods --- Local Outlier Factor ($LOF$) and Isolation Forest ($IF$). For $LOF$ we selected Minkowski ($LOF_{ED}$) and DTW ($LOF_{DTW}$) as metrics to refer the difference between ordinary and elastic metrics. We do not consider neural network based methods in the performance comparison, since they cannot be easily trained on small datasets with feasible detection performance. To our best knowledge, there is no similar anomaly detection approach to the proposed \edtwa~method which we can compare with. 
    
    For evaluation purposes, we opted for datasets from three different domains to prove a broader relevance of our method. Specifically: a) healthcare domain represented by ECG Heartbeat Categorization Dataset~\citep{kachuee2018ecg}\footnote{https://www.kaggle.com/shayanfazeli/heartbeat}, b) Industry domain by CNC Milling Machines Vibration Dataset~\citep{tnani2022smart}\footnote{ https://github.com/boschresearch/CNC\_Machining}, and c) Public services by NYC taxi passengers dataset from Numenta Anomaly Benchmark~\citep{lavin2015evaluating}\footnote{https://github.com/numenta/NAB}.
    
    The aim of the evaluation is to compare the ability to identify anomalous series of the \edtwa~method with selected methods. In experiment with ECG Heartbeat Categorization Dataset, we used expert knowledge to verify the normality of selected normal patterns. In experiments with other two datasets, no expert knowledge was applied.
    
    Evaluation of anomaly detection methods is also a challenging task. Two general criteria might be applied: 1. number of true anomalies detected, 2. number of false anomalies detected, i.e., well-known trade-off between precision and recall balance~\citep{elmrabit2020evaluation}. It is hard to evaluate any detection model without more context, especially in the nature of anomaly detection where datasets have usually strongly imbalanced classes.
    That is why, in each experiment, we report F1 score as well as accuracy due to strongly imbalanced datasets --- anomalies appear disproportionately less in the datasets.
    As is usual in anomaly detection, we denote anomalous samples in any dataset as positive class and normal samples as negative class.
    
    In the next subsections, we describe proposed baseline method for comparison, selected evaluation datasets and the results comparing selected methods with \edtwa.
    
    \subsubsection{Baseline method: Generalised DTW based anomaly detection}
    The baseline method is built on DTW normalised distance. In the training phase of the base method, DTW distances of representative normal pattern $R$ and all training time series $\mathcal{D} = \{D(R,T) | T \in \mathcal{T}\}$ are calculated. Based on $\mathcal{D}$, we define anomalous threshold $\theta_{base}$ as maximum DTW distance, i.e., $\theta_{base} = \max (\mathcal{D})$. If $D(R,Q)$ exceeds $\theta_{base}$ during detection of unseen time series $Q$, we report the time series $Q$ as anomalous. 
    
    Undoubtedly, there are pathological warping path cases which satisfy normalised distance with unexpected behaviour. These cases are not correctly detected by this naive method.
        
    \subsubsection{Healthcare: ECG Heartbeat Categorization}    
\begin{figure*}
            \centering
            \includegraphics[width=1.0\textwidth]{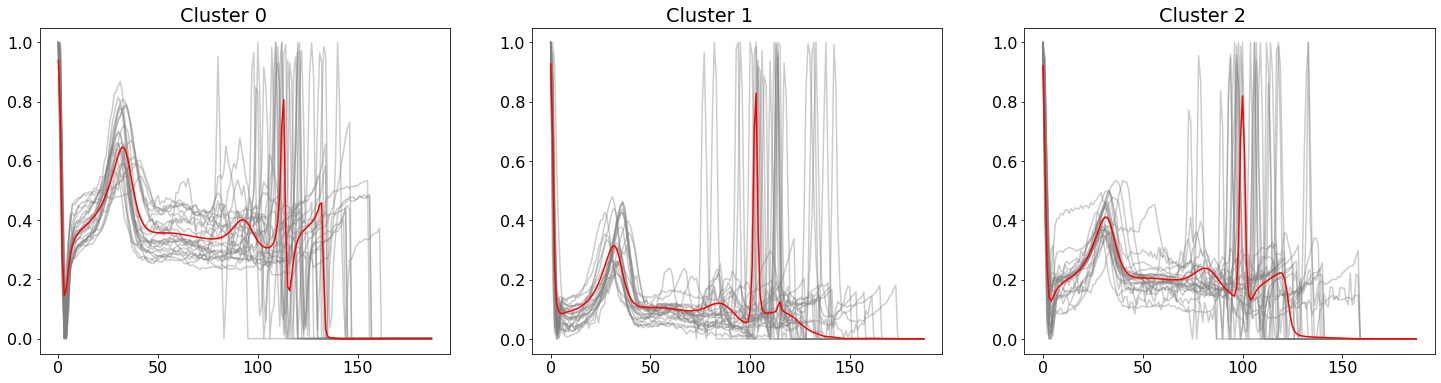}
            \caption{ECG samples clustered into 3 clusters using k-Means and DTW distance according to Elbow method. Light grey series are random ECG samples within given cluster, red series is softdtw barycenter averaging calculated normal representative.}
            \label{fig:evaluation_dataset_clusters}
        \end{figure*}

        The dataset consists of 4046 normal and 10506 anomalous samples. All the samples were cropped, downsampled and padded with zeroes if necessary to the fixed length of 188 by the author of the original dataset. To make the task easier for the expert (cardiologist), we applied simple preprocessing and provide initial normal behaviour patterns (healthy ECG signal examples). Firstly, we estimated an optimal number of ECG clusters within normal samples by Elbow method with k-Means clustering. We identified 3 clusters. Secondly, in order to express the most realistic normal ECG representative for each cluster, we applied DTW barycenter averaging (DBA) technique~\citep{petitjean2011global}. Figure \ref{fig:evaluation_dataset_clusters} shows clusters with their representative normal behaviour patterns. Calculated representatives were verified by cardiologist, who confirmed they represent the normal ECG patterns. The verified representative patterns were set as ground-truth normal patterns for given dataset.
        
        The aim of this experiment was to evaluate our method with regard to its accuracy and F1-score while using normal patterns verified by the expert. Given ECG dataset, we split normal ECG samples into train and test sets by 70 to 30. For testing purposes, we selected at random 900 normal samples out of 1335 available and 100 abnormal samples out of 10506 samples to make testing dataset reflecting real-life situations where anomalies occur much less often than normal data. 
        
        For $DTW_{base}$, we applied data-driven approach to calculate maximum allowed normalised warping path distance threshold as 9.04. Isolation Forrest best detection rate was achieved by trained estimators' size 250. Local Outlier Factor with Minkowski and DTW distances achieved based detection rates with neighbours count set to 30 and 400 respectively.
        
        \edtwa~method has detection window size $l$ parameter that must be set prior using it. To estimate the best possible results, we applied window hyper parameter tuning with initially estimated value window size 5. The highest accuracy (\textbf{92.30\%}) was achieved with window size 60.
        
        Table \ref{table:evaluation_results_healthcare}. summarises detailed results in the context of all selected methods and the key evaluation metrics. Our method and $LOF_{DTW}$ accomplished nearly identical results speaking in terms of the F1 score and Accuracy. Based on F1 score it is evident that even with imbalanced classes $LOF_{DTW}$ and \edtwa~achieved significantly better and consistent results than the other methods. The difference between $LOF_{DTW}$ and \edtwa~is in the number of FP and FN. $LOF_{DTW}$ detects more anomalies and therefore the number of FP is naturally higher. On the other hand, \edtwa~produces less false alarms and detects less anomalies.
        
        Local Outlier Factor configurations provide interesting results indirectly proving that elastic measures are able to reach higher detection rate score with improved partial results over standard measures too. The disadvantage of $LOF_{DTW}$ is its high computational complexity and hence long detection time. Depending on configuration, $LOF_{DTW}$ might need internally more $DTW$ computations than \edtwa~to achieve the same results. Our method needs to compute at most N times $DTW$ (number of normal patterns) while $LOF_{DTW}$ for the desired number of neighbours.
        
        \begin{table}
            \centering
            \caption{Anomaly detection results on ECG dataset for selected methods covering F1-score and accuracy. $LOF_{DTW}$ and \edtwa~achieve significantly better results than the other three methods.}
            \label{table:evaluation_results_healthcare}
            \begingroup
            \renewcommand{\arraystretch}{1.4}
            \begin{tabular}{l|rrrrcc}
                \textbf{Method} & TN   &  FP &  FN & TP & F1                 & Accuracy           \\ \hline
                $DTW_{base}$    & 837  &  63 &  65 & 35 & 0.3535             & 0.8720             \\
                IF              & 693  & 207 &  50 & 50 & 0.2801             & 0.7430             \\
                $LOF_{mink}$    & 752  & 148 &  79 & 21 & 0.1561             & 0.7730             \\
                $LOF_{DTW}$     & 839  &  61 &  21 & 79 & \textbf{0.6583}    & \underline{0.9180} \\
                \edtwa          & 860  &  40 &  38 & 62 & \underline{0.6139} & \textbf{0.9220}    \\
            \end{tabular}
            \endgroup
        \end{table}
        
\begin{figure*}[!ht]
            \centering
            \begin{subfigure}[t]{0.5\linewidth}
                \centering
                \includegraphics[width=0.99\linewidth]{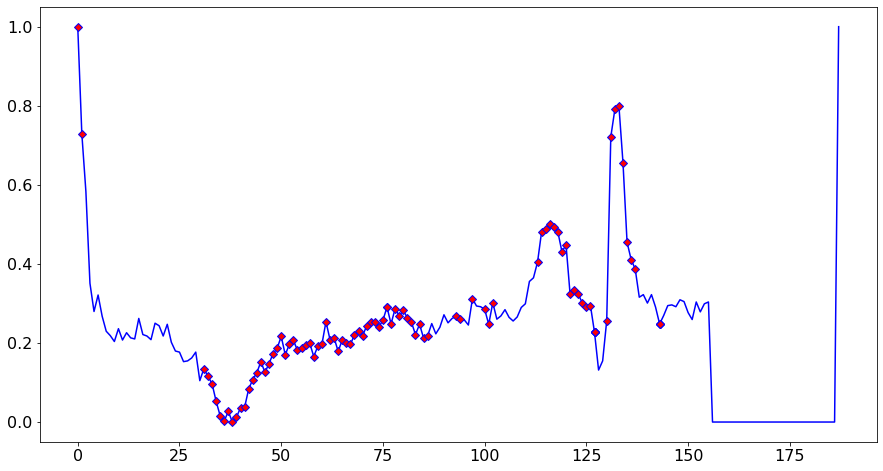}
            \end{subfigure}%
            \begin{subfigure}[t]{0.5\linewidth}
                \centering
                \includegraphics[width=0.99\linewidth]{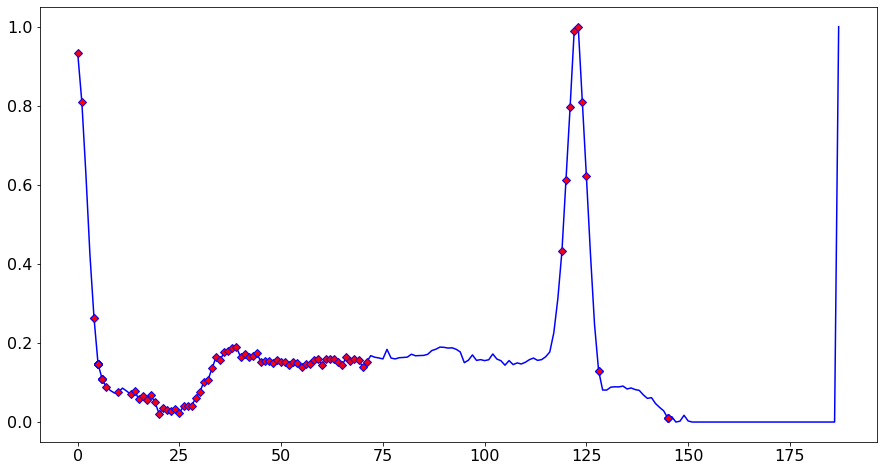}
            \end{subfigure}%
            
            \begin{subfigure}[t]{0.5\linewidth}
                \centering
                \includegraphics[width=0.99\linewidth]{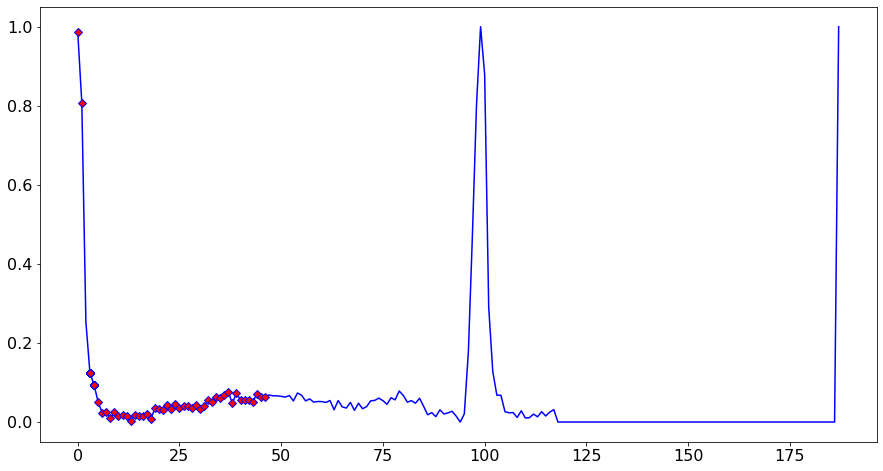}
            \end{subfigure}%
            \begin{subfigure}[t]{0.5\linewidth}
                \centering
                \includegraphics[width=0.99\linewidth]{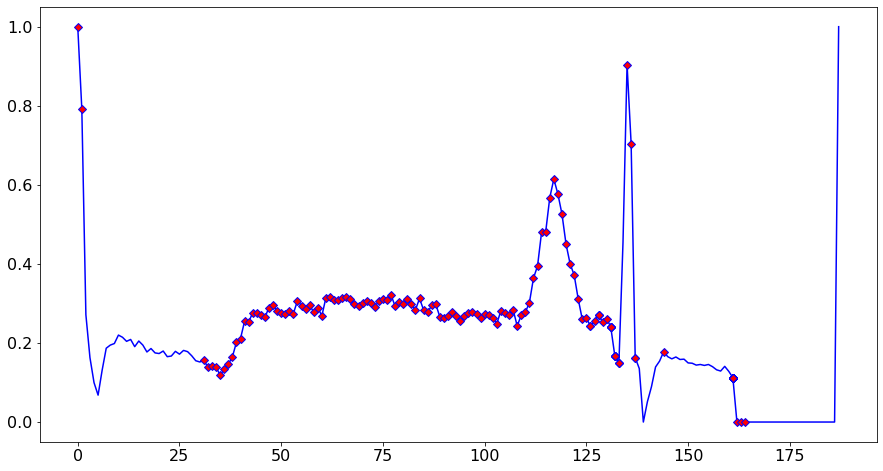}
            \end{subfigure}%
            
            \caption{Examples of detected anomalous ECG samples with marked anomalous data points (red markers).}
            \label{fig:edtwa_ecg_anomalous_samples}
        \end{figure*}

        The benefit of provided expert knowledge in the form of normal ECG patterns undoubtedly lies in lower computational complexity while achieving close results to $LOF_{DTW}$ method. Our method needed to compute internally $DTW$ at most 3 times while $LOF_{DTW}$ at most 400 times. Figure~\ref{fig:edtwa_ecg_anomalous_samples} shows examples of detected anomalous ECG samples with marked suspicious points for each time series.
        
    \subsubsection{Industry: CNC Milling Machines}
        The dataset represents real-world industrial vibration data collected from a brownfield CNC milling machine. Vibrations were measured as acceleration using a tri-axial accelerometer (Bosch CISS Sensor). Each recording is captured by 3 timeseries for X-, Y- and Z-axes. The data was collected in 4 different timeframes, each lasting 5 months from February 2019 until August 2021, and labelled. The data consists of three different CNC milling machines each executing 15 processes. Table \ref{table:evaluation_cnc_dataset_description} covers dataset properties from the process perspective such as number of normal and anomalous series occurrences and their lengths.
        
        \begin{table}
            \centering
            \caption{CNC Milling machines dataset properties overview. Each row describes different process from the dataset.}
            \label{table:evaluation_cnc_dataset_description}
            \begingroup
                \renewcommand{\arraystretch}{1.4}
                \begin{tabular}{l|rr|rr}
                    \multirow{2}{*}{\textbf{Process}} &
                      \multicolumn{2}{c|}{\textbf{\# Series}} &
                      \multicolumn{2}{c}{\textbf{Series Length}} \\
                     &
                      \multicolumn{1}{c}{\textbf{Normal}} &
                      \multicolumn{1}{c|}{\textbf{Anomalous}} &
                      \multicolumn{1}{l}{\textbf{Average}} &
                      \multicolumn{1}{l}{\textbf{Stdev}} \\ \hline
                    OP00 & 83  & 1  & 267698 & 6237  \\
                    OP01 & 136 & 7  & 57617  & 4408  \\
                    OP02 & 148 & 4  & 87400  & 3815  \\
                    OP03 & 68  & 2  & 163427 & 9999  \\
                    OP04 & 105 & 7  & 130704 & 10892 \\
                    OP05 & 114 & 6  & 41847  & 3860  \\
                    OP06 & 84  & 4  & 183887 & 9031  \\
                    OP07 & 148 & 10 & 48734  & 3601  \\
                    OP08 & 112 & 7  & 75063  & 3572  \\
                    OP09 & 113 & 1  & 210575 & 4564  \\
                    OP10 & 112 & 7  & 94789  & 4544  \\
                    OP11 & 68  & 6  & 116717 & 4729  \\
                    OP12 & 118 & 5  & 95226  & 3769  \\
                    OP13 & 142 & 0  & 66462  & 1907  \\
                    OP14 & 81  & 3  & 66688  & 3401 
                \end{tabular}
            \endgroup
        \end{table}
        
        The aim of the experiment is to evaluate the performance of the methods while solving task of anomalous processes identification. In this experiment we considered no human expert.
        
        Time series with normal label were split into train - test by 70 to 30 percent. All anomalous time series were kept in test dataset, i.e., 1142 normal time series in training set, 490 in test set and 70 anomalous time series in test dataset. For \edtwa, we applied the same process like in the previous task --- clustered timeseries using k-Means and calculated normal behaviour patterns using soft-DTW barycenter averaging.
        
        Table \ref{table:evaluation_results_cnc} presents the results of the experiment in detail. Our method achieved the second best accuracy result (93.88\%) while $LOF_{DTW}$ achieved 95.02\%. However, while comparing F1 score, our method clearly outperformed the other methods. $LOF$ based methods are able to safely detect normal time series, but true anomalies detection comparing with the \edtwa~is not performing well. Most of the anomalies were in this case missed.
        
        \begin{table}
            \centering
            \caption{Anomaly detection results for CNC machines dataset.}
            \label{table:evaluation_results_cnc}
            \begingroup
            \setlength{\tabcolsep}{3pt} 
            \renewcommand{\arraystretch}{1.4}
            \begin{tabular}{l|rrrrrrr}
            \multicolumn{1}{l|}{\textbf{Method}} &
              \multicolumn{1}{c}{TN} &
              \multicolumn{1}{c}{FP} &
              \multicolumn{1}{c}{FN} &
              \multicolumn{1}{c}{TP} &
              \multicolumn{1}{c}{F1} &
              \multicolumn{1}{c}{Accuracy} \\ \hline
            $DTW_{base}$ & 460 &  37 & 25 &   1 &            0.0313  &            0.8815  \\
            \emph{IF}    & 347 & 150 & 20 &   6 &            0.0659  &            0.6749  \\
            $LOF_{mink}$ & 485 &  12 & 25 &   1 &            0.0512  &            0.9292  \\
            $LOF_{DTW}$  & 496 &   1 & 25 &   1 & \underline{0.0714} &    \textbf{0.9502} \\
            \edtwa       & 472 &  25 &  7 &  19 &    \textbf{0.5428} & \underline{0.9388}
            \end{tabular}
            \endgroup
        \end{table}

    \subsubsection{Public Services: NYC Taxi Passengers}
        \label{sec:evaluation_nyc}
        The dataset contains aggregated count of passengers served by NYC taxi services for every 30 minutes between the $1^{st}$~July~2014 and the~$31^{st}$~January~2015. The dataset contains known anomalies such as the NYC marathon, Thanksgiving, Christmas, New Years Day, and a snow storm around the $27^{th}$~January~2015. In our case, the training dataset contains in our case date range from 2014-07-01 till 2014-11-01 while the rest of days is used for testing purposes. All anomalous days are kept in testing dataset in order to keep train to test datasets in relatively expected sizes and still respect logical order of the days - it should not be possible to train model on future days and evaluate on the past ones. Table~\ref{table:evaluation_nyc_anomalous_dates} contains a list of all identified anomalous dates.
        
        \begin{table}
            \centering
            \caption{Identified anomalous days with the description}
            \label{table:evaluation_nyc_anomalous_dates}
            \begingroup
                \renewcommand{\arraystretch}{1.1}   
                \begin{tabular}{l|l|c}
                    \renewcommand{\arraystretch}{1.1}
                    \textbf{Date} & \textbf{Anomalous Event} & \textbf{Time Series Shape}                                 \\ \hline
                    2014-11-01    & NYC Marathon             & \includegraphics[height=20px]{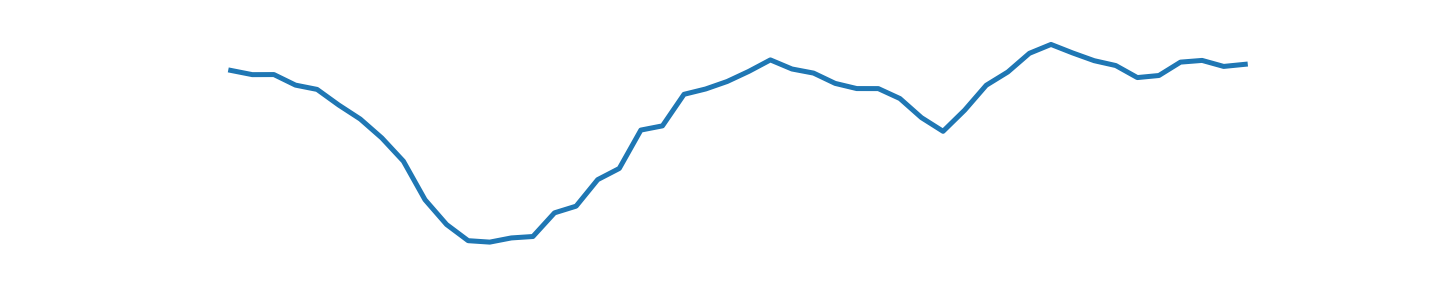} \\
                    2014-11-02    & NYC Marathon             & \includegraphics[height=20px]{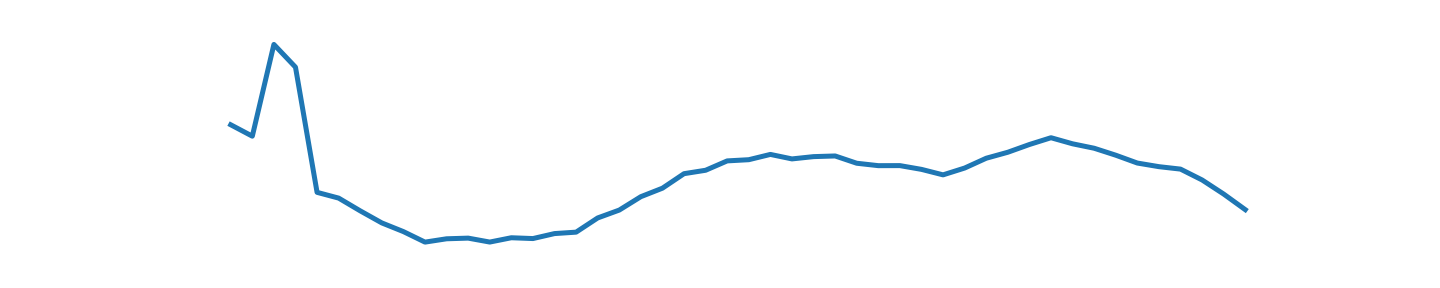} \\ 
                    2014-11-27    & Thanksgiving             & \includegraphics[height=20px]{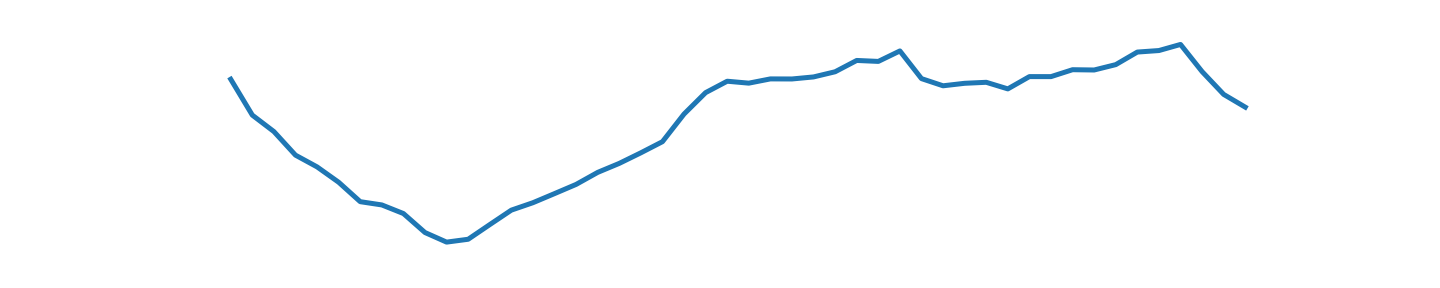} \\ 
                    2014-12-24    & Christmas                & \includegraphics[height=20px]{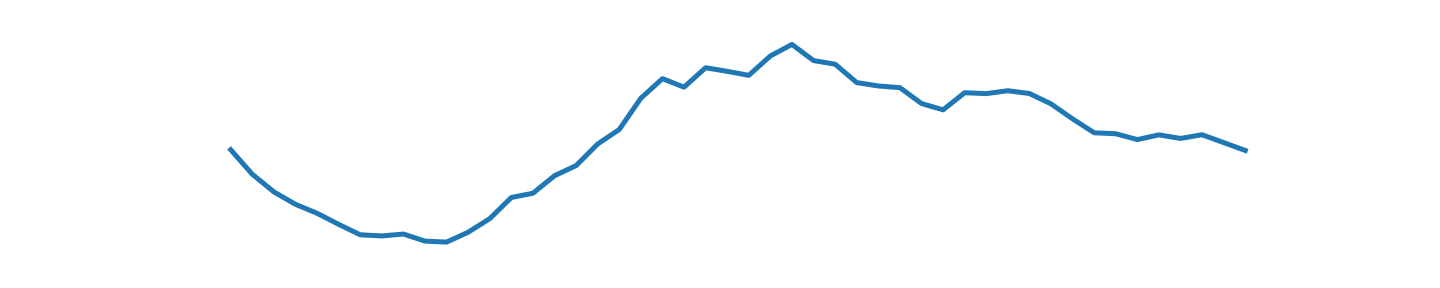} \\
                    2014-12-25    & Christmas                & \includegraphics[height=20px]{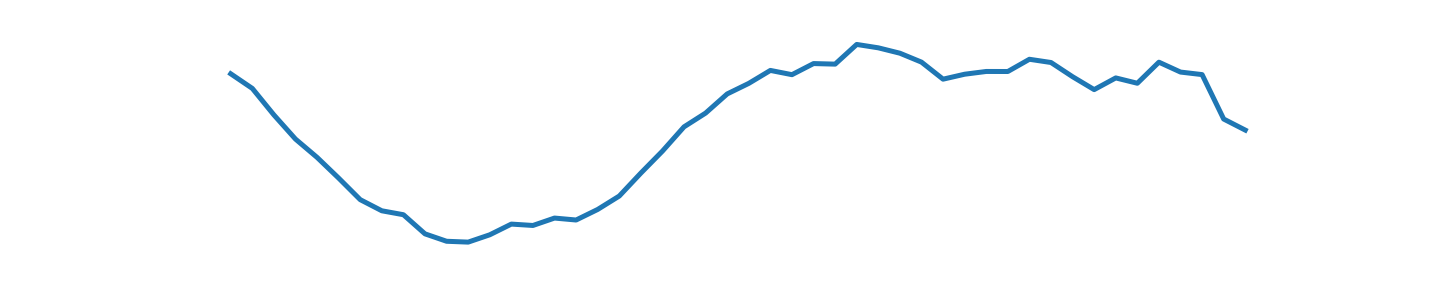} \\
                    2014-12-26    & Christmas                & \includegraphics[height=20px]{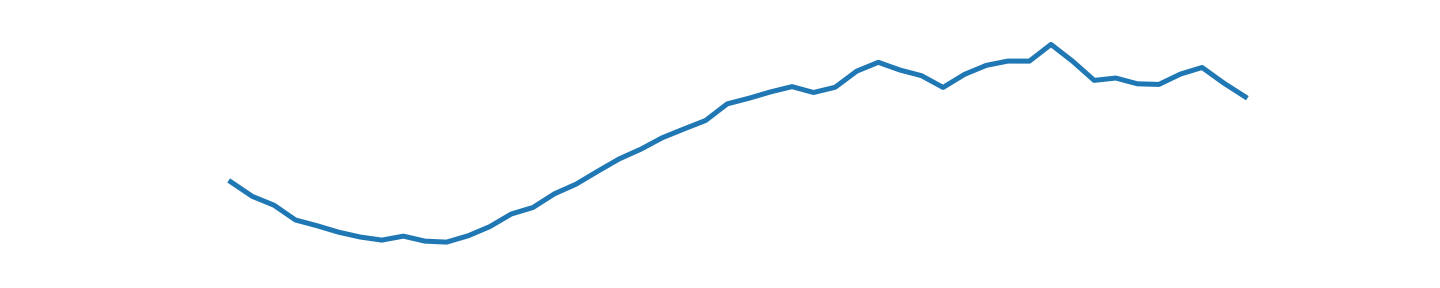} \\       
                    2014-12-31    & New Years Eve            & \includegraphics[height=20px]{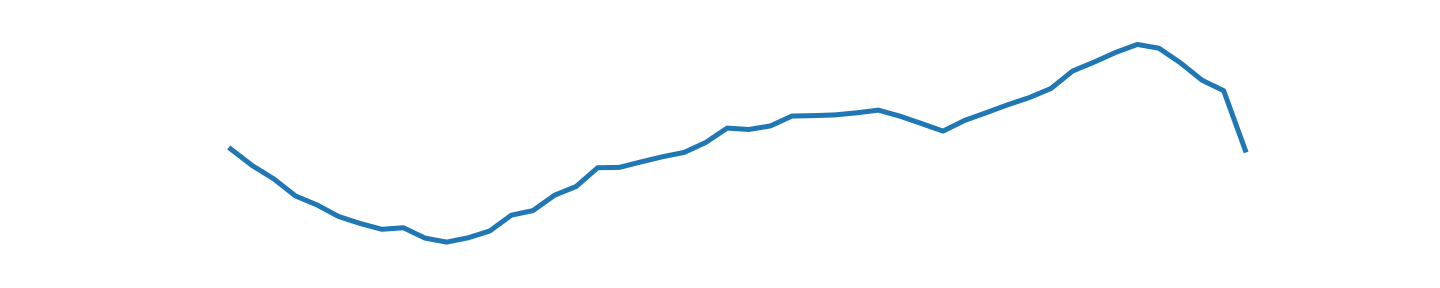} \\
                    2015-01-01    & New Years day            & \includegraphics[height=20px]{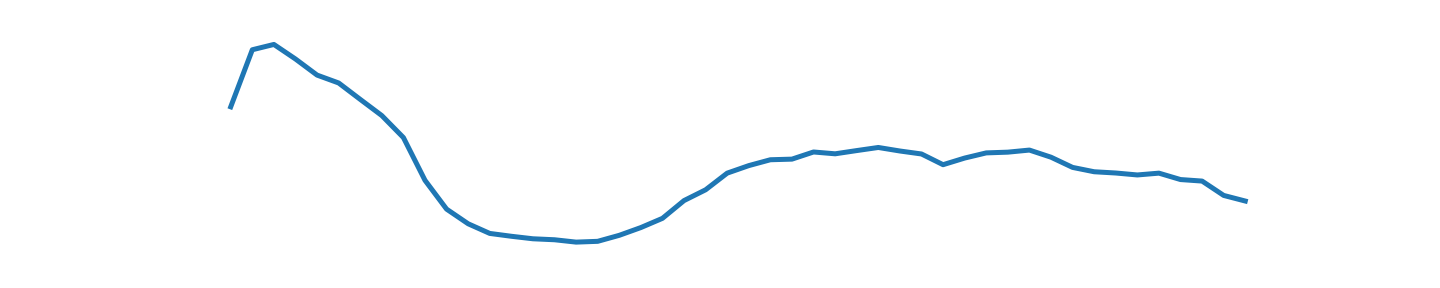} \\
                    2015-01-27    & Snow Storm               & \includegraphics[height=20px]{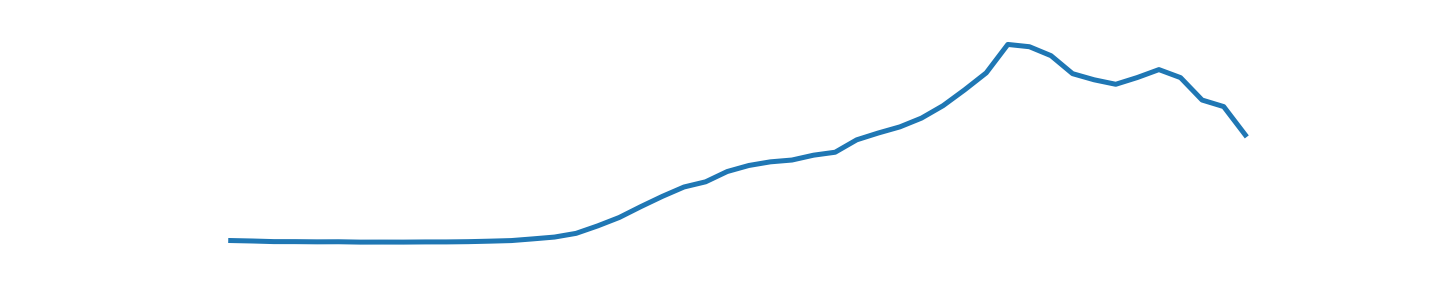} 
                \end{tabular}
            \endgroup
        \end{table}
        
        The aim of this experiment was to evaluate the performance of the methods while solving task of anomalous days identification.
        
        In this task, we focused purely on the detection aspect and did not embrace expert knowledge. 
        For \edtwa, we applied the same process like in the previous experiment --- clustered timeseries using k-Means and calculated normal behaviour patterns using soft-DTW barycenter averaging.
        
        Table \ref{table:evaluation_results_nyc}. provides detailed results in the context of all selected methods and key evaluation metrics. Our method detected correctly 8 of 10 anomalous days while incorrectly identified 4 normal days as anomalous, i.e., F1 score \emph{0.7273} and accuracy \emph{0.9368}. Isolation Forrest identified the most anomalous days, but with the highest false alarms penalty - 35 alarms. Both $LOF$ methods achieved the same F1 score and Accuracy. The same results are due to no significant shift in passengers' behaviour.
        
        \begin{table}
            \centering
            \caption{Anomaly detection results for NYC passengers dataset.}
            \label{table:evaluation_results_nyc}
            \begingroup
            \setlength{\tabcolsep}{3pt} 
            \renewcommand{\arraystretch}{1.4}
            \begin{tabular}{l|rrrrrrr}
            \multicolumn{1}{l|}{\textbf{Method}} &
              \multicolumn{1}{c}{TN} &
              \multicolumn{1}{c}{FP} &
              \multicolumn{1}{c}{FN} &
              \multicolumn{1}{c}{TP} &
              \multicolumn{1}{c}{F1} &
              \multicolumn{1}{c}{Accuracy} \\ \hline
            $DTW_{base}$ & 81 &  4 &  6 &  4 & 0.4444 & 0.8947 \\
            \emph{IF}    & 50 & 35 &  1 &  9 & 0.3333 & 0.6211 \\
            $LOF_{mink}$ & 81 &  4 &  4 &  6 & 0.6000 & 0.9158 \\
            $LOF_{DTW}$  & 81 &  4 &  4 &  6 & 0.6000 & 0.9158 \\
            \edtwa       & 81 &  4 &  2 &  8 & 0.7273 & 0.9368 
            \end{tabular}
            \endgroup
        \end{table}

        The performance of proposed \edtwa~method was compared to four anomaly detection state-of-the-art methods on three datasets. The \edtwa~method achieved the highest accuracy on healthcare dataset \textbf{$92.30\%$} while the second best method ($LOF_{DTW}$) achieved accuracy $91.80\%$. In the second experiment, related to industry dataset, our method detected the largest number of true anomalies (19 of 26). The second best method in this case was $IF$, which detects 6 out of 26 anomalies. In the third experiment from the transportation domain, our method correctly identified 8 out of 10 anomalous days. $IF$ detected 9 of 10 days, but with significantly higher FP (35 vs 4 in our case).
    
        To show the importance of using elastic measures, $LOF$ with $ED$ and $DTW$ measures were compared on all datasets. NYC dataset contains time series without any distortions, thus both methods gained the same scores. However, the results for both accuracy and F1 score evaluated on other 2 datasets prove our premise that for datasets that inherently contain time series with distortions, methods with elastic measures such $DTW$ achieve higher detection rate over those using $ED$ (see Tables \ref{table:evaluation_results_healthcare} and \ref{table:evaluation_results_cnc}). 
        
        \subsection{Evaluation of HITL Contribution}
        The benefits of HITL concept are apparent in the situations, when it is not possible to learn the properties of the problem sufficiently well from the training data, namely when the training dataset is corrupted or too small. We demonstrate the proposed application of HITL in our model in the experiment with small training dataset containing first 200 samples of \emph{ECG Heartbeat Categorization} dataset. Trained \edtwa~ model with HITL evaluated 1000 test samples. The expert was simulated by the procedure that labelled samples that the model is not sure about by known ground truth labels.
        
        When applying HITL in decision making process, it is important not to overwhelm the expert with enormous number of queries. In our solution, the expert was asked to label time series with which the method is uncertain, i.e., time series $t$ for which $0.25$ and $0.30$ holds. Out of 1000 test samples, the expert was asked to label 84 samples, i.e., only $8.4\%$ of all test samples. When including queried samples, the F-measure improved from $0.139$ to $0.192$ what represents $38.12\%$ performance gain. Accuracy improved from $72.9\%$ to $85.7\%$.
        
        The application of HITL process did not only improve the decision of samples labelled by the expert but updating of the model after each sample labelling continuously retrained the model. And this process enabled the model to increase the number of correctly detected and not expert labelled samples by 44 (TN increased by 50 and TP decreased by 6), i.e., $4\%$, see Table \ref{table:evaluation_results_hitl}.
        
        The experiment demonstrated the ability of the model to incorporate the expert knowledge into the decision process while continuously improving the model and hence the overall process.
        
        \begin{table}
            \centering
            \caption{Anomaly detection results for \edtwa~method with and without HITL on ECG dataset.}
            \label{table:evaluation_results_hitl}
            \begingroup
            \setlength{\tabcolsep}{3pt} 
            \renewcommand{\arraystretch}{1.4}
            \begin{tabular}{l|rrrrrrr}
            \multicolumn{1}{l|}{\textbf{Method}} &
              \multicolumn{1}{c}{TN} &
              \multicolumn{1}{c}{FP} &
              \multicolumn{1}{c}{FN} &
              \multicolumn{1}{c}{TP} &
              \multicolumn{1}{c}{F1} &
              \multicolumn{1}{c}{Accuracy} \\ \hline
            \edtwa            & 707 & 193 & 78 & 22 & 0.139 & 0.729 \\
            $\edtwa_{EXPERT}$ & 840 &  60 & 83 & 17 & 0.192 & 0.857 
            \end{tabular}
            \endgroup
        \end{table}

\section{Conclusion and Future work}
    \label{sec:conclusion}

    Our main goal was to improve anomaly detection with emphasis on computational complexity, precision and human-in-the-loop aspects. We proposed anomaly detection method based on automated DTW paths knowledge extraction. It is based on stored deviations of training time series from representative normal pattern. In regard to high memory and computational complexity of the other evaluated methods, strength of our work lies in low constant memory requirements with the option to simultaneously train and update detection model. The experiments proved that with lower computational complexity, our method reached competitive results with state-of-the-art DTW-based methods. \edtwa~ can provide instant feedback from the expert back to detection model. We have experimentally showed that when dealing with small training dataset, our approach can exploit human-in-the-loop concept to improve the detection performance. 
    
    This study enhances our understanding of anomaly detection in the context of human-in-the-loop concept. To deepen our research, we plan to improve knowledge integration with our method to tackle model explainability and increase trust in automated detection model actions. Knowledge extracted from warping paths of normal data can be used to localize anomalous points. We plan to modify our method for online time series processing to make it exploitable in even more areas. Due to ability of continuous model updates, we plan to exploit this feature to tackle concept drift phenomenon with appropriate mechanism.
    
\section*{CRediT authorship contribution statement} 
\textbf{Matej Kloska}: Methodology, Software, Data curation, Investigation, Writing – original draft, review \& editing. \textbf{Gabriela Grmanová}: Validation, Supervision, Writing – review \& editing. \textbf{Viera Rozinajová}: Conceptualization, Supervision, Writing – review \& editing.
    
\section*{Declaration of Competing Interest}
The authors declare that they have no known competing financial interests or personal relationships that could have appeared to influence the work reported in this paper.
    
\section*{Acknowledgement}
This research was partially supported by TAILOR, a project funded by EU Horizon 2020 research and innovation programme under GA No. 952215 and by the project Life Defender - Protector of Life, ITMS code: 313010ASQ6, co-financed by the European Regional Development Fund (Operational Programme Integrated Infrastructure).

\bibliographystyle{elsarticle-harv} 
\bibliography{elsarticle}





\end{document}